\documentclass{article}
\usepackage{amssymb, amsmath, latexsym}
\usepackage{amsfonts, amsbsy}
\newtheorem{theorem}{Theorem}
\newtheorem{lemma}{Lemma}
\newtheorem{proposition}{Proposition}

\newtheorem{notation}{Notation}
\newtheorem{remark}{Remark}
\begin{document}
\title{\bf Study of all the periods of a Neuronal Recurrence Equation}
\author{ Serge Alain Eb\'{e}l\'{e}, Ren\'{e} Ndoundam \footnote{Corresponding author: ndoundam@yahoo.com} \\
{\small University  of Yaounde I, LIRIMA, Team  GRIMCAPE, P.o.Box  812 Yaounde, Cameroon}  \\
{\small CETIC, Yaounde, Cameroon}  \\
{\small IRD, UMI 209, UMMISCO, IRD France Nord, F-93143, Bondy, France;  }  \\
{\small Sorbonne Unversit\'es, Univ. Paris 06, UMI 209, UMMISCO, F-75005, Paris, France } \\
{\small E.mail : sergeebele@yahoo.fr, ndoundam@yahoo.com} 
             }
\date{}
\maketitle { }
\begin{abstract} We characterize the structure of the periods of a neuronal recurrence equation. Firstly,
we give a characterization of k-chains in 0-1 periodic sequences. Secondly, we characterize the periods of all cycles of
some neuronal recurrence equation. Thirdly, we explain how these results can be used to deduce the existence of the 
generalized period-halving bifurcation.
\end{abstract}
\section{Introduction}
The human brain can be viewed as a set of interconnected neurons. Caianiello \cite{Caia 66, CLR 67} suggested to model the brain 
using the following threshold automata network:
\begin{equation}      \label{LL:nnum3}
x_i(t+1) = {\bf 1} \Big( \sum_{j=1}^{n} \sum_{s=1}^{k} a_{ij}(s) x_{j}(t+1-s) - \theta_i \Big) \ \ \ \ 1 \ \leq i \ \leq n , \ \ t \geq k-1     
\end{equation}
where:
\begin{tabbing}
\hspace*{.25in}\=\hspace{2ex} \kill
\> $x_j(t+1-s)$ is the state of the neuron $j$ at time $t+1-s$,  \\
\> $a_{ij}(s)$ represents the influence of the neuron $j$ at time $t+1-s$ on the neuron $i$ at time $t+1$, \\
\> $\theta_i$ is the threshold of the excitation of the neuron $i$,  \\
\> $\sum_{j=1}^{n} \sum_{s=1}^{k} a_{ij}(s) x_{j}(t+1-s)$ is the potential of the neuron $i$ at time $t$,  \\
\> $n$ is the number of the neurons of the network,   \\
\> $k$ is the size of the memory,  \\
\> {\bf 1}[$u$] = 0 if $u \ < \ 0$, and {\bf 1}[$u$] = 1 if  $u \geq 0$. \\
\end{tabbing}

The dynamics of this model has been studied in some particular cases:

\begin{enumerate}
\item In the Equation (\ref{LL:nnum3}), when $k=1$, we obtain the following equation:
\begin{equation}      \label{LL:numy7}
x_i(t+1) = {\bf 1} \Big( \sum_{j=1}^{n} a_{ij} x_{j}(t) - \theta_i \Big) \ \ \ \ 1 \ \leq i \ \leq n    
\end{equation}
which models the dynamic behavior of $n$ interconnected
neurons of memory size $1$. These networks were introduced by McCulloch
and Pitts \cite{MP 43}, and are quite powerful.  
\item In the Equation (\ref{LL:nnum3}), when $n=1$, we obtain the following equation:
\begin{gather}   \label{eeq:n1}
x(n) = {\bf 1} \Biggl( \sum_{j=1}^{k}a_{j}x(n-j)- \theta \Biggl)
\end{gather}
introduced by Caianiello and De Luca \cite{CL 66} which models the
dynamic behavior of a single neuron with a memory, that does not
interact with other neurons. 
\end{enumerate}

Neural networks are usually implemented by using electronic components or
are simulated by a software on a digital computer. One way in which the collective
properties of a neural network may be used to implement a computational task is 
through the {\it energy minimization} concept. The Hopfield network is a well-known 
example of such an approach. It has attracted  a wide  attention 
in  literature as a {\it content-addressable memory} \cite{SH 99}.  \\
Caianiello networks have been studied by Goles \cite{G 85} and Ndoundam \cite{NT 00}. 
Many studies have been devoted to the McCulloch and Pitts' neural networks \cite{GM 90, GM 89, GO 81,G 81, Ma 95, Fog 87, SH 99}.
Matamala \cite{Ma 95} studied McCulloch and Pitts' reverberation neural networks (i.e. neural networks of McCulloch and 
Pitts where each state of the system, after a finite number of steps comes back to itself, hypercube permutation).  \\

Cosnard,  Moumida,  Goles  and T. de St. Pierre \cite{CMG 88}  showed  the following result in the case of palindromic memory:
\begin{proposition} \cite{CMG 88}
If the interacting coefficients  $(a_1, a_2, \cdots, a_k)$  verify
\[
a_i \ = \ a_{k+1-i}  \ \ \forall  i \in \mathbb{N}, \ 1 \leq i \leq k  
\]
Then the length  of each cycle is a divisor of  $k+1$.
\end{proposition}
$\blacksquare$      \\
In the case of  j-palindromic memory, they also showed:
\begin{proposition} \cite{CMG 88}
If the interacting coefficients  $(a_1, a_2, \cdots, a_k)$ are j-palindromic,  i.e. verify
\begin{itemize}
\item  $a_1  \ =  \ a_2 \ = \  \cdots \ a_{j} \ = \ 0$
\item  $a_i  \ = \ a_{k+j+1-i}  \ \ \ \ \forall  i \in \mathbb{N}, \  j+1  \leq i \leq  k$
\end{itemize}  
Then the length  of each cycle is a divisor of  $k+j+1$.
\end{proposition}
$\blacksquare$     \\
When the memory are geometric sequence, they showed the following result:
\begin{proposition} \cite{CMG 88}
If the interacting coefficients  verify  $a_i \ = \ -(b^i)$ with $b \in ]0, \frac{1}{2}]$
Then the length  of each cycle is  less or equal to $k+1$.
\end{proposition}
$\blacksquare$   \\
In the case of positive geometric sequence, they showed
\begin{proposition} \cite{CMG 88}
If the interacting coefficients  verify  $a_i \ = \ (b^i)$  with  $b \in ]0, \frac{1}{2}]$
Then the length  of each cycle is  $1$.
\end{proposition}
$\blacksquare$     \\
Another results have been established on neuronal recurrence equations modeling neurons with memory \cite{GM 90, CMG 88, 
CTT 92, CG 84, Cos 99, NM 00a, NM 00b, NT 04, TT 93, Mou 89}. From the point of view of the period:
\begin{itemize}
\item in \cite{CTT 92, TT 93, NM 00a, NM 00b, NT 04}, the authors {\bf didn't study all the cycles} generated by the 
       neuronal recurrence equation;
\item in this paper, we  {\bf are studying all the cycles} generated by the neuronal recurrence equation $\{ y(n)   \ : \ n  \geq 0 \}$.  
\end{itemize}
From the point of view of bifurcation:
\begin{itemize}
\item  in \cite{NT 03}, we studied the dynamics of the sequence $\{ z(n)   \ : \ n  \geq 0 \}$ from {\bf one and only one} initial configuration. We characterized {\bf only one} cycle of the sequence $\{ z(n)   \ : \ n  \geq 0 \}$; 
\item  in \cite{Ndou 12}, for any $d$ ( $0  \leq d  \leq   \rho(m)-1$ ),  we studied the dynamics of the sequence $\{ z(n, d)   \ : \ n  \geq 0 \}$ from {\bf one and only one} initial configuration. We characterized {\bf only one} cycle of the sequence $\{ z(n, d)   \ : \ n  \geq 0 \}$; 
\item  in this paper, for any $d$ ( $0  \leq d  \leq   \rho(m)-1$ ),  we are showing how to study the dynamics of the sequence $\{ z(n, d)   \ : \ n  \geq 0 \}$ from {\bf any} initial configuration. We are showing how to characterize the length of {\bf all} cycles of the sequence $\{ z(n, d)   \ : \ n  \geq 0 \}$. 
\end{itemize}
Our work, from some point of view, is similar to the work of Matamala \cite{Ma 95} in the sense that we study all the periods.

The paper is organized as follows: in Section 2, some previous results are presented. Section 3 presents 
a characterization of k-chains in 0-1 periodic sequences. Section 4 is devoted to the characterization 
of the period length of all the cycles. In section 5, we study a bifurcation. 
Concluding remarks are stated in Section 6.

\section{Previous  Results}

Given a finite neural network, the configuration assumed by the
system at time $t$ is ultimately periodic. As a consequence, there
is an integer $p \ > \ 0$ called the period (or the length of a
cycle) and another integer $T \geq 0$ called the transient length such that:
\begin{list}{\texttt{$\bullet$}}{}
\item $Y(p+T) = Y(T)$
\item $\nexists \ T' \ and \ p' \ \ (T', p') \ne (T, p) \ \ \ T \geq T' \ and
        \ p \geq p' \ \ {\rm such \ that} \ Y(p'+T') \ = \ Y(T')$
\end{list}
where $Y(t) = (x(t), x(t-1), \dots , x(t-k+2), x(t-k+1))$. The
period and the transient length of the sequences generated are good
measures of the complexity of the neuron. A bifurcation occurs when a small smooth change made to the 
parameter values (the bifurcation parameters) of a system, causes a sudden 'qualitative' or topological
 change in its behaviour. A period-halving bifurcation in a dynamic system, is a bifurcation 
 in which the system switches to a new behaviour with half the period of the original system from some initial 
 configuration. A generalized period-halving bifurcation is a period-halving bifurcation from any initial
 configurations. \\

Cosnard, Tchuente and Tindo \cite{CTT 92} show the following lemma:
\begin{lemma} \cite{CTT 92}   \label{elll:nn1} \\
If there is a neuronal recurrence equation with memory length $k$ that generates 
sequences of periods $p_1, p_2,  \dots , p_r$, then there is a neuronal recurrence
equation with memory length $kr$ that generates a sequence of period
$r \times lcm(p_1, \cdots , p_r)$.
\end{lemma}    
$\blacksquare$  \\
Lemma \ref{elll:nn1} does not take into account the study of
the transient length. One can amend Lemma \ref{elll:nn1} to obtain the following
lemma:

\begin{lemma}  \label{elnll:puon2}    \cite{Ndou 12}
If there is a neuronal recurrence equation with memory length $k$
that generates a sequence $\{x^{\jmath}(n) : n \geq 0 \} \ ,  1
\leq \jmath \leq g$ of transient length $T_{\jmath}$ and of period
$p_{\jmath}$, then there is a neuronal recurrence equation with
memory length $kg$ that generates a sequence of transient length
$g \times max(T_1, T_2, \dots , T_g)$ and of period $Per$. $Per$ is defined as follows: \\
{\it First case:} $\exists \ j, \ 1 \leq \ j  \ \leq \ g$ such that $p_j \geq 2$  \\
\[
Per \ = \ g  \times lcm(p_1, \cdots , p_g).
\]
{\it Second case:} $p_j = 1$ ;  $\forall \ j, \ 1 \leq \ j  \ \leq \ g$.  \\
\[
Per {\rm \ is \ a \ divisor \ of \ } g.
\]
\end{lemma}
$\blacksquare$  \\

Cosnard and Goles \cite{CG 84} studied the bifurcation in two particular case of neuronal recurrence
equations. \\
{\it Case \ 1: \ Geometric coefficients and bounded memory} \\
Cosnard  and Goles completely described the structure of the bifurcation of the following equation:
\[
x_{n+1} = {\bf 1} \Biggl( \theta - \sum_{i=0}^{k-1} b^i x_{n-i} \Biggl)
\]
when $\theta$ varies. They showed that the associated rotation number is an increasing number of the
parameter $\theta$.  \\
{\it Case \ 2: \ Geometric coefficients and unbounded memory.} \\

Cosnard  and Goles completely described the structure of the bifurcation of the following equation:
\[
x_{n+1} = {\bf 1} \Biggl( \theta - \sum_{i=0}^{n} b^i x_{n-i} \Biggl)
\]
when $\theta$ varies. They showed that the associated rotation number is a devil's staircase. \\

The next section is devoted to the study of k-chains.
\section{Characterization of k-chains in 0-1 periodic sequences}

We recall the concept of k-chains in 0-1 periodic sequences \cite{GM 90} which  is useful in the study of
the limit orbits. Let $Y \ = \ ( y(t) \ : \ t \in  \mathbb{N} )$ be a periodical sequence of 0's and 1's; suppose
 that the period $\gamma(Y)$ ( which is a priori unknown ) divides $T$. Thus $y(t)  \in \{ 0 \ , 1 \}$ for any 
 $t \in \mathbb{Z}$ and $y(t) = y(t')$ when $t \equiv t' \pmod{T}$. \\
In studying period lengths, we shall deal with sets invariant under translations\cite{GM 90}, so the following
notation will be useful: if $\Gamma  \subset \mathbb{Z}_T$, $l  \in \mathbb{Z}$, we write:
\[
\Gamma + l \  = \ \{ \ t+l \  ( \ mod \ T \ ) \ : \ t \ \in \ \Gamma \}
\]
Let us partition the set $\mathbb{Z}_T$ into $\Gamma^0(Y) \ = \ \{ \ t \ \in \ \mathbb{Z}_T \ : \ y(t) = 0 \}$ and
 $\Gamma^1(Y) \ = \ \{ \ t \ \in \ \mathbb{Z}_T \ : \ y(t) = 1 \}$ which is called the support of Y. The period of 
 the set $\Gamma^1(Y)$ is the smallest positive number $\gamma$ such that $\Gamma^1(Y) + \gamma \ = \ \Gamma^1(Y)$. The following 
result was established in \cite{GM 90} : the period of the sequence ( i.e. $\gamma(Y)$ ) is equal to the period of 
$\Gamma^1(Y)$. It is shown in \cite{GM 90} that:
\[
\gamma(Y) \ {\rm  divides \ k \ if \ and \ only \ if } \ \Gamma^1(Y) \ + k = \ \Gamma^1(Y)
\]
Now let us define k-chains ( for $k \geq 1 $ ) contained in the support $\Gamma^1(Y)$. A subset $C \subset \Gamma^1(Y)$ 
is called a k-chain if and only if it is of the form $C \ = \ \{ \ t+kl \ ( \ {\rm mod \ T } \ )\ : \ 0 \leq l \leq s-1 \}$ for some $s \geq 1$.
So a k-chain is a subset $C \ = \ \{ \ t+kl \ \in \ \mathbb{Z}_T \ : \ 0 \leq \ l \ \leq s-1 \  \}$ such that $y(t') \ = \ 1$ for any $t' \ \in \ C$.  \\

We characterize the 0-1 sequence which contains two different chains.
\begin{lemma}  \label{lem:mel1}    
{\bf If} a 0-1 sequence $\{ u(n) \ : \ n \geq 0 \}$ contains:
\begin{itemize}
\item an $\ell_1$-chain,
\item an $\ell_2$-chain,
\item such that $\ell_1$ and  $\ell_2$ are relatively prime.
\end{itemize}   
{\bf Then}  \\
$\exists \ t \ \in \mathbb{N}$ such that:
\begin{itemize}
\item $u(t) \ = \ 1$,
\item $u(t+\ell_1) \ = \ 1$,
\item $u(t+\ell_2) \ = \ 1$. 
\end{itemize}   
\end{lemma}
$\blacksquare$  \\

We use the lemma \ref{lem:mel1} to characterize all the periods of all the attractors.

\section{Characterization of the periods of all the cycles}

Let us consider a positive integer $m$ and a positive real number $\theta \geq \ 2m$, we note:
\begin{notation}
\begin{itemize}
\item $p_0, p_1, \cdots, p_{s-1}$ are prime numbers taken between $2m$ and $3m$ such that $p_i \ < \ p_{i+1}$,
\item $\alpha_i \ = \ 3m- p_i$, $0 \ \leq \ i \ \leq s-1$,
\item $k=6m$,
\end{itemize}
\end{notation}
and we define the coefficients as follows:
\begin{equation}  \label{ccch9:eq2}
coef_1(j) = \begin{cases}
         (\theta / 2) - \alpha_i     &\text{if $j \ = \ 3m- \alpha_i, \ 0 \leq \ i \ \leq s-1$;}  \\
         (\theta / 2) + \alpha_i     &\text{if $j \ = \ 2(3m-\alpha_i), \ 0 \leq \ i \ \leq s-1$;}  \\
         -k(\theta+m)           &\text{otherwise}.
      \end{cases}
\end{equation}

The coefficients defined in Equation (\ref{ccch9:eq2}) are analog to those defined in \cite{TT 93}.
For each $i$, $0 \leq i \leq s-1$, the first $k$ terms of the sequence
$\{ x^{\alpha_i}(n) \ : \ n \geq 0  \}$ are defined as follows:
\begin{equation}  \label{ch9:eq4}
x^{\alpha_i}(0) x^{\alpha_i}(1) \cdots x^{\alpha_i}(k-1) \ = \ \underbrace{00 \dots 0}_{2\alpha_i} \underbrace{100 \dots 0}_{3m-\alpha_i} \underbrace{100 \dots 0}_{3m-\alpha_i}
\end{equation}

$\forall \ n \ \geq \ k$, the term $x^{\alpha_i}(n)$ of the sequence $\{ x^{\alpha_i}(n) \ : \  n \in \mathbb{N} \}$
is defined as follows:
\[
x^{\alpha_i}(n) \ = \ {\bf 1}( \sum_{i=1}^{k} coef_1(i) x^{\alpha_i}(n-i) - \theta )
\]

By using the technique developped by Tchuente and Tindo \cite{TT 93}, it is easy to prove the following lemma:

\begin{lemma} \label{lleeme:mel22}     
The sequence $\{ x^{\alpha_i}(n) \ : \  n \in \mathbb{N} \}$ describes a cycle of length $3m-\alpha_i$ of the following 
form:
\[
\underbrace{00 \dots 0}_{2\alpha_i} \underbrace{100 \dots 0}_{3m-\alpha_i} 
\underbrace{100 \dots 0}_{3m-\alpha_i}\underbrace{100 \dots 0}_{3m-\alpha_i} \cdots  
\underbrace{100 \dots 0}_{3m-\alpha_i}  \cdots
\]
\end{lemma}
$\blacksquare$  \\
In \cite{TT 93}, the authors {\bf didn't study all the cycles} generated by the neuronal recurrence equation.
One of our aims is {\bf to study all the cycles} generated by some neuronal recurrence equation. \\

We construct the sequence $\{ u(n) \ : \ n \geq 0 \}$ generated by the neuronal recurrence equation 
\begin{equation}  \label{ch99:eq66}
u(n) \ =  \ {\bf 1} \Bigl( \sum_{j=1}^{k} coef_1(j) u(n-j) - \theta \Bigl)  \ , \ n \geq k
\end{equation}
such that the initial terms are defined as follows:
\begin{equation}  \label{ch999:eq1}
u(0)u(1) \cdots u(k-1) \ \in \{0, 1 \}^k.  
\end{equation}

Let us characterize the period of the sequence $\{ u(n) \ : \ n \geq 0 \}$ by showing
the following proposition:

\begin{proposition}  \label{prop:prop25}
The sequence $\{ u(n) \ : \ n \geq 0 \}$ converges
\begin{itemize}  
  \item to the null  sequence i.e.  to  $0 \ 0 \cdots 0 \ 0 \cdots 0 \ 0 \cdots$ \ , \ or   
  \item to one of the sequences $\{ x^{\alpha_i}(n) : n \ \geq \ 0 \} \ , \ 0 \ \leq \ i  \ \leq \ s-1$.   
\end{itemize}
\end{proposition}
$\blacksquare$      \\    
 {\bf Example:}  In the aim to give an idea of the basin of attraction of the sequence $\{ u(n) \ : \ n \geq 0  \}$,
 we choose the following parameters:
 \[
 m = 5, \ \theta = 12, \ p_0 = 11, \ p_1 = 13, \ {\rm and} \ k = 30.  
 \]
We build from the preceding parameters the following neuronal recurrence equation
\begin{equation}   \label{kk0:kk8}
u(n) \ =  \ {\bf 1} \Bigl( \sum_{j=1}^{30} coef_1(j) u(n-j) - \theta \Bigl)  \ , \ n \geq 30
\end{equation}
where:
\begin{equation}   \label{kk5:kk9}
      coef_1(j) =
      \begin{cases}
      2   \ ,   &\text{if $j = 11$ } \\
      4   \ ,   &\text{if $j = 13$ } \\
      10  \ ,  &\text{if  $j = 22$ } \\
      8   \ ,  &\text{if  $j = 26$ } \\
      -510 \ ,   &\text{otherwise.} 
      \end{cases}
\end{equation}
Let us note:
\begin{multline}  \notag
config(i) \ = \ \{\, u(0)u(1) \cdots u(28)u(29) \ : \  \forall i, \ 0 \leq i \leq 29 \ , \ u(i) \in \{ 0, 1 \}  \\
                   \rm{and \ the \ neuronal \ recurrence \ equation \ defined \ by \ equation} (\ref{kk0:kk8}) \ {\rm from \ the}  \\
                   {\rm initial \ terms} \ u(0)u(1) \cdots u(28)u(29) \ {\rm converges \ to \ a \ cycle \ of \ length} \ i \,\}. 
\end{multline}
We also note $\chi(i) \ = \ card( config(i) )$. By numerical simulations, the values of the sequence
 $\chi(i)$ are:
\begin{align}
 &\chi(i)   = 0  \ \ if \ i \ \notin \ \{1, \ 11, \ 13 \}     \notag   \\
 &\chi(1)   = 1073713157     \notag     \\
 &\chi(11)  = 4094           \notag     \\
 &\chi(13)  = 24573.          \notag    
\end{align}
  
\begin{notation} 
Let us define the memory length of some neuronal recurrence equations as follows:
\[
h \ = \ s \times k = 6m \times s.
\]
\end{notation}

Let $\{ y(n)  : n \geq 0 \}$ be the sequence whose first $h$ terms are defined as follows: 
\begin{equation} \label{eeq:n9}
y\left( i \right) \ \in \ \{ 0 , 1 \} , \ \  0 \leq  i \leq -1+h,
\end{equation}
and the other terms are generated by the following neuronal recurrence
equation:
\begin{equation}  \label{eeq:n10}
y(n) = {\bf 1} \left[ \sum_{f=1}^{h} coef_2(f) y(n-f)- \theta_1 \right]  \ ; \ n \geq h
\end{equation}
where
\begin{equation}   \label{eeq:n11}
      coef_2(f) =
      \begin{cases}
      coef_1(j) \ ,   &\text{if $f = s \times j, \ \ 1 \leq j \leq k$ } \\
      0 \ ,   &\text{otherwise.} 
      \end{cases}
\end{equation}
\begin{equation} \label{eeq:n12}
\theta_1 = \theta.
\end{equation}
The parameters $coef_1(j)$ are those defined in Equation (\ref{ccch9:eq2}).
\begin{remark} \label{r:n1}    
(a) The first $h$ terms of the sequence $\{ y(n) : n \geq 0 \}$
are obtained by taking any element of the set $\{ 0, 1 \}^h$. \\
(b) The coefficients $coef_2(f)$ of neuronal recurrence Equation (\ref{eeq:n10}) are obtained 
by applying the construction of Lemma \ref{elll:nn1} to the parameters defined by 
Equation (\ref{ccch9:eq2}). \\
\end{remark}
Our aim is to characterize the structure of all the periods of the sequence $y(n)$ from
a qualitative point of view. The next theorem gives the period of the sequences $\{ y(n) : n \geq 0 \}$.
\begin{theorem}  \label{ell:n7}
From any initial term, the sequence $\{ y(n) \ : \ n \geq 0 \}$ converges to a cycle of length :
\begin{itemize}
\item  $s \times lcm( elt_1 , elt_2, \dots , elt_s )$ where $elt_i \ \in \  \{ p_0 , p_1, \dots, p_{s-1} \}$
for any $i \in \{ 0, 1, 2, \dots, s-1 \}$, \ or 
\item $p$ where $p$ is equal to $1$.
\end{itemize}
\end{theorem} 
$\blacksquare$ \\

In the next section, we show how to apply the previous technique to the study of bifurcation of the neuronal recurrence
equation $z(n,d)$.

\section{Generalized Bifurcation of the neuronal recurrence equation}

Let us define the neuronal recurrence equation $\{ \tilde{x}(n) \ : \ n \ \geq \ 0 \ \}$ by the following recurrence:
\begin{equation} \label{ch9:eq6}
\tilde{x}(t) =  {\bf 1} \Bigl( \sum_{j=1}^{k_2} coef_3(j) \tilde{x}(t-j) - \bar{\theta} \Bigl) \ ; \  t \geq k_2
\end{equation}
where $coef_3(j)$ is defined as follows:  \\

{\bf First case:} $s$ is even and $\forall \ i_2 \in \mathbb{N}, \ 0 \leq i_2 \leq s-1$

\begin{equation}   \label{eeeqq:n2}
      coef_3(j)  =
      \begin{cases}
 2 &\text{if  $j \in R1(\alpha_{i_2}) \ {\rm and} \  j \leq  \frac{3 \times s \times p_{i_2}}{2} \ ,$}  \\
-2 &\text{if $j \in R1(\alpha_{i_2}) \ {\rm and} \ j \ >  \frac{3 \times s \times p_{i_2}}{2} \ , $}  \\
-4 \times k_2  &\text{otherwise. }
      \end{cases}
\end{equation}

{\bf Second case:} $s$ is odd, $s \geq 3$ and $\forall \ i_2 \in \mathbb{N}, \ 0 \leq i_2 \leq s-1$ 

\begin{equation}   \label{eeq:n333}
      coef_3(j)  =
      \begin{cases}
 2 &\text{if $j \in R1(\alpha_{i_2}) \ {\rm and} \ j \leq  \frac{(3 \times s)-1}{2} \times p_{i_2} \ , $ } \\
-2  &\text{if $j \in R1(\alpha_{i_2}) \ {\rm and} \ \frac{(3 \times s )+1}{2} \times p_{i_2}  \leq j \leq (2s -2) \times p_{i_2}, $} \\
-1 &\text{if $j \in \{(2s-1) \times p_{i_2} \ , \ 2s \times p_{i_2} \}$ \ , }   \\
-4  \times  k_2  &\text{otherwise. }
      \end{cases}
\end{equation}

The parameters $R1(\alpha_i)$, $\bar{\theta}$ and $k_2$ are defined as follows:

\begin{align}
 R1(\alpha_i) &= \{ jp_i : j = 1, \dots , 2s \}  \label{pp1:nn1}   \\
       &= \{ p_i, 2p_i, \dots , (-1+ 2s)p_i , 2 \times s \times p_i \}, 
     \ \ 0 \ \leq i \ \leq -1+s;   \label{pp1:nn2}   \\
 R2 &= \{ i : i = 1, \dots, k_2 \} = \{ 1, 2, \dots , -1+k_2, k_2 \};  \label{pp1:nn3}    \\
 R3 &=   \bigcup_{i=0}^{-1+s}  R1( \alpha_i);    \label{pp1:nn4}     \\
 R4 &=  R2 \setminus R3 ;     \label{pp1:nn5}     \\
\bar{\theta} &= \ 2 \times s ; \label{pp1:nn6}  \\
 k_2 &= (6m-1) \times s.     \label{pp1:nn7} 
\end{align}

By applying the technique developped in sections 2 and 3, and the one developped in 
 \cite{Ndou 12} to the neuronal recurrence equation defined by Equation (\ref{ch9:eq6}), it is easy to  
construct a family of neuronal recurrence equations  $\{ z(n, d) : n \geq 0 \}$ which verify the following theorem.

\begin{theorem}  \label{theo:numb20}
$\forall m, d \in \mathbb{N}$ such that $m \geq e^2$ and $0 \leq d \leq s-2$, we construct
a set of neuronal recurrence equations whose behaviour has the following characteristics:
\begin{itemize}
\item  From any initial configuration, the neuronal recurrence equation $\{ z(n, d) : n \geq 0 \}$
 converges to a cycle  of  length $s \times lcm( elt_1 , elt_2, \dots , elt_s )$ where 
 $elt_i \ \in \  \{ p_{d+1} , p_{d+2}, \dots, p_{s-1} \}$ for any $i \in \{ 1, 2, \dots, s \}$, or to a cycle of length 1.
\item From any initial configuration, the neuronal recurrence equation $\{ z(n, s-1) : n \geq 0 \}$ converges to a fixed
      point (i.e. the period of a cycle is 1). 
\end{itemize}
\end{theorem}
$\blacksquare$  \\

In other words, the first part of Theorem \ref{theo:numb20} can be interpreted as follows: in some cases, the
length of the cycles of the neuronal recurrence equation $\{ z(n, d-1) : n \geq 0 \}$ is divided by $p_d$ to obtain
the length of cycles of the neuronal recurrence equation $\{ z(n, d) : n \geq 0 \}$.    \\
By perturbation, we can build the neuronal recurrence equation $\{ z(n, d) : n \geq 0 \}$ from the neuronal recurrence 
equation $\{ z(n, d-1) : n \geq 0 \}$. 

\begin{remark} \label{rrur:vv9} 
The new contributions in this paper with respect to the previous works are: \\
Firstly, from the point of view of the period:
\begin{itemize}
\item in \cite{CTT 92, TT 93, NM 00a, NM 00b, NT 04}, the authors {\bf didn't study all the cycles} generated by the 
       neuronal recurrence equation;
\item in this paper, we  {\bf studied all the cycles} generated by the neuronal recurrence equation $\{ y(n)   \ : \ n  \geq 0 \}$.  
\end{itemize}
Secondly, from the point of view of bifurcation:
\begin{itemize}
\item  in \cite{NT 03}, we studied the dynamics of the sequence $\{ z(n)   \ : \ n  \geq 0 \}$ from {\bf one and only one} initial configuration. We characterized {\bf only one} cycle of the sequence $\{ z(n)   \ : \ n  \geq 0 \}$; 
\item  in \cite{Ndou 12}, for any $d$ ( $0  \leq d  \leq   \rho(m)-1$ ),  we studied the dynamics of the sequence $\{ z(n, d)   \ : \ n  \geq 0 \}$ from {\bf one and only one} initial configuration. We characterized {\bf only one} cycle of the sequence $\{ z(n, d)   \ : \ n  \geq 0 \}$; 
\item  in this paper, for any $d$ ( $0  \leq d  \leq   \rho(m)-1$ ),  we studied the dynamics of the sequence $\{ z(n, d)   \ : \ n  \geq 0 \}$ from {\bf any} initial configurations. We characterized the length of {\bf all} cycles of the sequence $\{ z(n, d)   \ : \ n  \geq 0 \}$. 
\end{itemize}
\end{remark}

\section{Conclusion} 
We have given a characterization of k-chains in 0-1 periodic sequences. This characterization allows us to determine
the periods of all cycles of some neuronal recurrence equations. From the structure of the periods of all cycles, we show 
how to build the family of neuronal recurrence equation $\{z(n,d) : n \geq 0 \}$ which admits a generalized period-halving 
bifurcation. The structure of the configuration of neuronal recurrence equation can be used
in steganography (see Second Approach and Third Approach of the paper \cite{NE 15}).\\
{\bf  Acknowledgements} \\
We thank the reviewers for their suggestions.This work was supported by {\it UMMISCO },  {\it  LIRIMA } and the
{\it University of Yaounde 1}.  \\

\appendix{\bf Appendix}   \\

{\bf Proof of Lemma \ref{lem:mel1}} \\
From the hypothesis, the sequence $\{ u(n) \ : \ n \geq 0 \}$ contains two chains, we can deduce that:
\begin{itemize}
\item $\exists \ a \ \in \mathbb{N} \ , \ 0 \leq a \ < \ \ell_1$
\item $\exists \ b \ \in \mathbb{N} \ , \ 0 \leq b \ < \ \ell_2$
\item $u(a+(i \times \ell_1)) \ = \ 1 \ , \forall i \in \mathbb{N}$
\item $u(b+(j \times \ell_2)) \ = \ 1 \ , \forall j \in \mathbb{N}$ 
\end{itemize} 
By hypothesis, integers $\ell_1$ and $\ell_2$ are relatively prime and from the definition of greatest 
common divisor, we can deduce:
\begin{equation}  \label{yy:ryy22}
\exists \ n_1 , n_2 \in \mathbb{Z} \ {\rm such \ that \ } n_1 \times \ell_1 + n_2 \times \ell_2 \ = \ 1
\end{equation}
From Equation (\ref{yy:ryy22}), we can easily deduce:
\begin{subequations}  \label{EEE:ggp}
 \begin{gather} 
    n_1 \times (b-a) \times \ell_1 + n_2 \times (b-a) \times \ell_2 \ = \ b-a     \label{EEE:ggp1}    \\         
    a + ( n_1 \times (b-a) \times \ell_1 ) = b - ( n_2 \times (b-a) \times \ell_2 )  \label{EEE:ggp2}       
 \end{gather}
\end{subequations}
From Equation (\ref{EEE:ggp2}),  it follows that $\exists \ i_0 , \ j_0 \ \in \mathbb{N}$ defined as follows:
\begin{subequations}  \label{EE:gppp}
 \begin{gather} 
i_0 \ = \ n_1 \times (b-a) + \Bigl( \bigl(1 + \mid n_1(b-a) \mid + \mid n_2(b-a) \mid \bigl) \times \ell_2 \Bigl)  \label{EE:gppp1}  \\         
j_0 \ = \ -n_2 \times (b-a) + \Bigl( \bigl(1 + \mid n_1(b-a) \mid + \mid n_2(b-a) \mid \bigl) \times \ell_1 \Bigl)  \label{EE:gppp2}   
 \end{gather}
\end{subequations}
such that:
\[
a + ( i_0 \times \ell_1 ) \ = \ b + ( j_0 \times \ell_2 )
\]
It suffices to choose $t = a + (i_0 \times \ell_1 )$.  \\
$\blacksquare$ \\

{\bf Proof of Proposition \ref{prop:prop25}}  \\
Without loss of generality, let us choose the following initial terms 
\[
u(0)u(1)u(2) \cdots u(k-1) \ \in \ \{ 0, 1 \}^k. 
\]
We suppose that from the following initial terms
\[
u(0)u(1)u(2) \cdots u(k-1) \ \in \ \{ 0, 1 \}^k 
\]
the sequence $\{ u(n) \ : \ n \geq 0 \}$ describes a transient of length $T_1$ and a cycle of length $P_1$.
We define the sequence $\{ w(n) \ : \ n \geq 0 \}$ as follows:
\[
\forall \ n \in \mathbb{N} \ \ \ w(n) = u(n+T_1). 
\]
In others words, the sequence $\{ u(n) \ : \ n \geq 0 \}$ converges to the attractor $\{ w(n) \ : \ n \geq 0 \}$.
The proof is divided into two parts: \\
Firstly, let us suppose that the sequence $\{ w(n) \ : \ n \geq 0 \}$ is not equal to one of the following
two sequences:
\begin{itemize}  
  \item to the null  sequence i.e.  to  $0 \ 0 \cdots 0 \ 0 \ 0 \cdots$  
  \item to one of the sequences $\{ x^{\alpha_i}(n) : n \ \geq \ 0 \} \ , \ 0 \ \leq \ i  \ \leq \ s-1$.   
\end{itemize}
We can extract from the sequence $\{ w(n) \ : \ n \geq 0 \}$ an $\ell$-chain such that :
\begin{equation}  \label{ret:ret25}
\ell \ \not\equiv  \ 0 \bmod{p_i} \ , \ 0 \ \leq i \ \leq s-1.
\end{equation}
Without loss of generality, let us assume that:
\begin{equation}  \label{ret55:ret255}
w(t_1) \ = \ 1 
\end{equation}
from the fact that the sequence $\{ w(n) \ : \ n \geq 0 \}$ admits an $\ell$-chain, we can deduce that:
\begin{equation}  \label{ret66:ret266}
w(t_1+\ell) \ = \ 1. 
\end{equation}

From Equation (\ref{ret:ret25}), we can easily deduce that:
\begin{equation}
coef_1(\ell) \ = \  -k(\theta+m).
\end{equation}
From the fact that:
\begin{itemize}
\item $w(\ell+t_1) \ = \ {\bf 1}( \sum_{j=1}^{k} coef_1(j) w(\ell+t_1-j) - \theta)$,
\item $coef_1(\ell) \ = \ -k(\theta+m)$,
\item $w(t_1) \ = \ 1$.
\end{itemize}
We deduce that : $w(\ell+t_1) \ = \ 0$. It follows that we have a contradiction with Equation (\ref{ret66:ret266}).
We can deduce that there is no $\ell$-chain in the sequence  $\{ w(n) \ : \ n \geq 0 \}$ that verifies Equation (\ref{ret:ret25}). \\

Secondly, let us suppose that on the sequence $\{ u1(n) \ : \ n \geq 0 \}$, there exist at least two 
different chains. \\

Without loss of generality, let us suppose that there exist on the sequence $\{ w(n) \ : \ n \geq 0 \}$ :
\begin{itemize}
  \item an $\ell_1$-chain such  that $\ell_1 \ = \ p_{i1} \ , \  0 \ \leq i1 \ \leq s-1$;
  \item an $\ell_2$-chain such  that $\ell_2 \ = \ p_{i2} \ , \  0 \ \leq i2 \ \leq s-1$;
  \item $l_1 \ < \ l_2$, i.e. $p_{i1} \ < \ p{i2}$. 
\end{itemize}

From the fact that the sequence $\{ w(n) \ : \ n \geq 0 \}$ admits two chains : $\ell_1$-chain and $\ell_2$-chain, we 
deduce by application of Lemma \ref{lem:mel1} that there exist $t_1 \in \mathbb{N}$ which verifies:

\begin{subequations}   \label{Eg:gp}
   \begin{gather}		
       w( t_1 ) = 1;            \label{Eg:ggp11}  \\
       w( t_1 + \ell_1 ) = 1;   \label{Eg:gp12}  \\
       w( t_1 + \ell_2 ) = 1.   \label{Eg:gp13}
  \end{gather}
\end{subequations}

We have : $2 \times m  \leq p_{i1} \ < \ p_{i2} \leq 3 \times m$. It follows that:
\begin{equation} \label{ret:ret35}
\imath \ = \ \ell_2 - \ell_1 \ = \ p_{i2} - p_{i1} \ \leq \ m.
\end{equation}

From Equation (\ref{ccch9:eq2}) and Equation (\ref{ret:ret35}), we deduce that:

\begin{equation} \label{ret:ret37}
coef_1(\imath) \ = -k(\theta+m)
\end{equation}

Based on the facts that:
\begin{itemize}
  \item $w( t_1 + \ell_1 ) \ = \ 1$, 
  \item $coef_1(\imath) \ = \ -k(\theta+m)$,
  \item $w( t_1 + \ell_2 ) \ = \ {\bf 1} \Bigl( \sum_{j=1}^{k} coef_1(j)  w(t_1+\ell_2-j) - \theta \Bigl)$.
\end{itemize}
We deduce easily that $w( t_1 + \ell_2 ) = 0$. This is a contradiction with the Equation (\ref{Eg:gp13}).
We easily deduce that the sequence $\{ w(n) \ : \ n \geq 0 \}$ contains one and only one chain. \\
$\blacksquare$  \\

{\bf Proof of Theorem \ref{ell:n7}}  \\
Based on Lemma \ref{elnll:puon2} and Proposition \ref{prop:prop25}, we deduce the result.  \\
$\blacksquare$  \\


\begin{thebibliography}{2}
\bibitem{GM 90} E. Goles and S. Mart\'{\i}nez,
{``Neural and Automata Networks, Dynamical Behaviour and Applications"}
(Norwell, MA, Kluwer Academic Publishers, 1990)
\bibitem{SH 99} S. Haykin , {``Neural Networks : A Comprehensive Foundations''}
(Upper Saddle River, New Jersey : Prentice-Hall, 1999)
\bibitem{Caia 66} E. R. Caianiello, {``Decision Equations and Reverberations,"} {\it Kybernetic}, {\bf 3}(1966) 33-40
\bibitem{CLR 67} E. R. Caianiello and A. De Luca, L. Riccardi
{``Reverberations and Control of Neural Networks,"} {\it Kybernetic}, {\bf 4}(1967)
\bibitem{CL 66} E. R. Caianiello and A. De Luca,
{``Decision Equation for Binary Systems : Applications to Neuronal
Behavior,"} {\it Kybernetic}, {\bf 3}(1966) 33-40
\bibitem{CMG 88} M. Cosnard, D. Moumida, E. Goles and T.de.St. Pierre,
{``Dynamical Behavior of a Neural Automaton with Memory,"}  {\it Complex
Systems,} {\bf 2}, (1988), 161-176.
\bibitem{CTT 92} M. Cosnard, M. Tchuente and G. Tindo, {``Sequences
Generated by Neuronal Automata with Memory"},  {\it Complex Systems,} {\bf
6},1992,  13-20.
\bibitem{CG 84} M. Cosnard and E. Goles, {``Dynamique d'un automate \`a m\'emoire mod\'elisant le fonctionnement d'un neurone"},
{\it Comptes Rendus de l'Acad\'emie des Sciences}, {\bf tome 299, S\'erie I}, n°10,  459-461, 1984.
\bibitem{MP 43} W. S. McCulloch and W. Pitts, {``A Logical Calculus of the
Ideas Immanent in Nervous Activity"}, {\it Bulletin of
Mathematical Biophysics,} {\bf 5} (1943) 115-133.
\bibitem{Fog 87} F. Fogelman Soulie et al. {``Automata Networks and Artificial Intelligence"} in {\it Automata 
 Networks in Computers Science: Theory and Applications,} edited by F. Fogelman, Y. Robert, and M. Tchuente
 ( Manchester University Press, Manchester, 1987).
\bibitem{GO 81} E. Goles, J. Olivos,{``Comportement P\'eriodique des Fonctions \`a seuil
 Binaires et Applications"},  {\it Disc. App. Math.,} {\bf 3}, 1981, 95-105.
\bibitem{G 81} E. Goles, {``Fixed Point Behaviour of Threshold Functions on a finite set"},  {\it SIAM J. on Alg. 
 and Disc Methods}, {\bf 3(4)}, 1982, pages 529-531.
 \bibitem{G 85} E. Goles, {``Dynamical Behaviour of Neural Networks"},  {\it SIAM J. Disc. Alg. Methods}, {\bf 6}, 
 1985, pages 749-754.
\bibitem{GM 89} E. Goles and  S. Mart\'{\i}nez,{``Exponential Transient Classes of symmetric Neural networks for Synchronous
and Sequential Updating"},  {\it Complex Systems,} {\bf 3(1989)} 589-597.
\bibitem{Ma 95} M. Matamala,{``Recursive Construction of Periodic Steady State for Neural Networks"},  
{\it Theoretical Computer Science,} {\bf 143(2)} (1995) 251-267. 
\bibitem{NM 00a} R. Ndoundam and M. Matamala, {``Cyclic Evolution of Neuronal
Automata with Memory When All the Weighting Coefficients are Strictly
Positive"}, {\it Complex Systems}, {\bf 12(2000)}, 379-390.
\bibitem{NM 00b} R. Ndoundam and M. Matamala, {``No Polynomial Bound for the
Period of Neuronal Automata with Memory With  Inhibitory
Memory"}, {\it Complex Systems}, {\bf 12(2000)}, 391-397.
\bibitem{NT 00} R. Ndoundam and M. Tchuente, {``Cycles Exponentiels des Reseaux de Caianiello et 
 Compteurs en Arithmetique Redondante"}, {\it Techniques et Sciences Informatique}, {\bf Vol. 19},
 n°7/2000, pp 985-1008.
\bibitem{NT 03} R. Ndoundam and M. Tchuente, {``Exponential transient length generated by a neuronal 
 recurrence equation"}, {\it Theoretical Computer Science}, {\bf 306(2003)}, 513-533.
\bibitem{NT 04} R. Ndoundam and M. Tchuente, {``Exponential Period of Neuronal 
 Recurrence Automata with Excitatory Memory"}, {\it Complex Systems}, {\bf 15(2004)}, 79-88.
\bibitem{Ndou 12} R. Ndoundam, {``Period-Halving Bifurcation of a Neuronal Recurrence Equation"}, 
{\it Complex Systems}, {\bf 20(2012)}, 325-349.
\bibitem{NE 15} R. Ndoundam, S.G.R. Ekodeck, {``Pdf Steganography based on Chinese Remainder Theorem"}, 
{\it arXiv 1506.01256}, submitted to Journal of Information Security and Applications.
\bibitem{TT 93} M. Tchuente and G. Tindo, {``Suites g\'en\'er\'ees par une
\'equation  neuronale \`a m\'emoire"},{\it Comptes Rendus de
l'Acad\'emie des Sciences}, {\bf tome 317, Serie I},  625-630,
1993.
\bibitem{Cos 99} M. Cosnard, {``Dynamic Properties of An
Automaton with Memory"}, pages 293-302 in {\it  Cellular Automata : A Parallel
Model}, edited by M. Delorme and J. Mazoyer (Kluwer  Academics Publishers,
1999).
\bibitem{Mou 89} D. Moumida, {``Contribution \`a l'\'etude de la Dynamique
d'un Automate  \`a M\'emoire,"} ( Doctoral dissertation, University of
Grenoble, 1989).
\end{thebibliography}
\end{document}